%% file: main2.tex
\documentclass[compsoc,conference,a4paper,10pt,times]{IEEEtran}
\usepackage{booktabs}
\usepackage[T1]{fontenc} 
\usepackage{amsmath}
\usepackage{rotating}
\usepackage{epsfig,endnotes}
\usepackage[colorinlistoftodos,prependcaption,textsize=tiny]{todonotes}
\usepackage{url}

\usepackage{mathtools}
\usepackage[noend]{algpseudocode}
\usepackage{tabularx,colortbl}
\usepackage{multirow}
\usepackage[Symbolsmallscale]{upgreek}
\usepackage{algorithm}
\usepackage{algpseudocode}
\usepackage{mathtools}
\usepackage{tablefootnote}
\usepackage{siunitx}
\usepackage{listings}
\usepackage{enumitem}
\usepackage{color} 
\usepackage{footmisc}
\usepackage{amsmath}
\usepackage{hyperref}
\usepackage{booktabs}
\usepackage{subcaption}

\usepackage{mathtools, nccmath}
\usepackage{titlesec}

\title{Seeds Don't Lie: An Adaptive Watermarking Framework \\ for Computer Vision Models}

\newcommand\Mark[1]{\textsuperscript#1}

\begingroup
\centering
\author{
Jacob Shams\Mark{1}, Ben Nassi\Mark{1}, Ikuya Morikawa\Mark{2}, Toshiya Shimizu\Mark{2}, Asaf Shabtai\Mark{1}, Yuval Elovici\Mark{1}\\
\Mark{1}Ben-Gurion University of the Negev,
\Mark{2}Fujitsu\\
\{jacobsh, nassib\}@post.bgu.ac.il\\
\{morikawa.ikuya, shimizu.toshiya\}@fujitsu.com\\
\{shabtaia, elovici\}@bgu.ac.il
}
\endgroup

\makeatletter
\newcommand{\linebreakand}{%
  \end{@IEEEauthorhalign}
  \hfill\mbox{}\par
  \mbox{}\hfill\begin{@IEEEauthorhalign}
}
\makeatother

\begin{document}
\maketitle

\input{abstract}

\input{intro}

\input{related-works}
\input{hypothesis}
\input{method}

\input{evaluation}
\input{conclusion}

{\small
\bibliography{bibliography}
\bibliographystyle{abbrv}
}

\pagebreak
\input{appendix}

\end{document}

%% file: abstract.tex
\begin{abstract}
In recent years, various watermarking methods were suggested to detect computer vision models obtained illegitimately from their owners, however they fail to demonstrate satisfactory robustness against model extraction attacks \cite{lukas2021sok}.
In this paper, we present an adaptive framework to watermark a protected model, leveraging the unique behavior present in the model due to a unique random seed initialized during the model training. This watermark is used to detect extracted models, which have the same unique behavior, indicating an unauthorized usage of the protected model's intellectual property (IP).
First, we show how an initial seed for random number generation as part of model training produces distinct characteristics in the model's decision boundaries, which are inherited by extracted models and present in their decision boundaries, but aren't present in non-extracted models trained on the same data-set with a different seed.
Based on our findings, we suggest the Robust Adaptive Watermarking (RAW) Framework, which utilizes the unique behavior present in the protected and extracted models to generate a watermark key-set and verification model.
We show that the framework is robust to (1) unseen model extraction attacks, and (2) extracted models which undergo a blurring method (e.g., weight pruning).
We evaluate the framework's robustness against a naive attacker (unaware that the model is watermarked), and an informed attacker (who employs blurring strategies to remove watermarked behavior from an extracted model), and achieve outstanding (\textit{i.e.,} >0.9) AUC values.
Finally, we show that the framework is robust to model extraction attacks with different structure and/or architecture than the protected model.
\end{abstract}

%% file: intro.tex
\section {Introduction}
Deep Neural Networks (DNNs) are state-of-the-art machine learning models utilized in a variety of computer vision tasks, including image classification, object detection, and facial recognition~\cite{CHAI2021100134}. 
A DNN's design and development process is expensive, requiring substantial resources and efforts to produce a high-quality model (e.g., data collection, long training time, high level of computing capabilities, etc.), motivating the consideration of DNNs as the Intellectual Property (IP) of their owners. 
Such IP is an attractive target for actors with the incentive to steal or copy the model in order to benefit from the work of those who trained/own the model. 
Therefore, a mechanism enabling ownership verification is a necessity for protecting the IP rights of a DNN model owner and to prevent unauthorized distribution of the model.

In order to protect a DNN model and prevent unauthorized distribution of the model and/or related IP, \emph{DNN Watermarking} was suggested as a method of proof of ownership over a protected model.
The objective of DNN Watermarking is to embed/identify unique characteristics/behavior in a protected model, and to attest to the presence of these characteristics/behavior in other "suspect" models. Identifying such unique characteristics/behavior in another model would indicate the model was extracted from the protected model, potentially violating the rights of the protected model's owners.


Previous DNN Watermarking methods suggested static methods which rely on white-box access to the suspect model (e.g., ~\cite{nagai2018digital, fan2019extended}) and dynamic methods which rely on black-box access to the suspect model (e.g., ~\cite{merrer2019frontier,guo2018embedded,zhang2018protecting,li2019blind,namba2019robust,Zhao2020AFAAF,jia2021entangled,lukas2021deep,adi2018weakness}). These methods are capable of successfully identifying a DNN based on watermarked structural/behavioral characteristics of the model. However, these methods largely fail to take into account model extraction attacks and robust watermark blurring methods, which are used by attackers to illegitimately obtain DNN models and remove unique behavior attested to by the watermark, respectively \cite{lukas2021sok}.

In this paper, we show how an initial random seed used for random number generation as part of the model training creates unique behavior in the trained model that is inherited by extracted models.
Based on our findings, we propose the Robust Adaptive Watermarking (RAW) Framework, an adaptive watermarking framework that leverages the unique behavior present in a protected model and its extracted models to generate a watermark key-set and verification model.
We evaluate the robustness of the verification model, trained to detect extracted models based on query responses to the watermark key-set, against a naive attacker (unaware that the model is watermarked), and against an informed attacker (who employs blurring strategies to remove watermarked behavior from an extracted model).
In addition, we demonstrate that the unique behavior attested to by the verification model is transferred to extracted models possessing different model structure and/or architecture than the protected model, such that the RAW Framework generalizes to different structures and architectures.

The significance of our method is as follows:
    \textbf{(1) Robustness:} Previous watermarking methods were found to be vulnerable to unseen model extraction attacks, as well as informed (extraction + blurring) attacks.
    We perform an evaluation of our framework against model extraction attacks and demonstrate the framework’s robustness to (1) unseen extraction attacks, and (2) blurring attacks applied to an extracted model. We leverage an inherent property of the protected model, the initial seed utilized for random number generation when training the model, and show that this property produces unique behavior in the model that can't be easily removed by the attacker.
    \textbf{(2) Adaptive:} Most previous watermarking methods are rigid and don't take attacks into account when generating the watermark; once an attack is developed to counter the method, it becomes obsolete.
    Our proposed framework takes attacks into account in order to optimize watermark generation against a considered risk defined by the DNN owner, possessing a mechanism for adaptivity and customization such that the verification model can be updated and enhanced in response to new and improved attacks.
    \textbf{(3) Generality:} Our proposed method utilizes unique behaviors and characteristics inherent in the protected model and which transfer to extracted models, including those which possess different model structures and architectures.

%% file: related-works.tex
\section{Background \& Related Work}
\label{sec:related-works}

DNN Watermarking as a dedicated field began around 2018 with initial research performed by \cite{nagai2018digital} and \cite{merrer2019frontier}. 
Methods in this field can be divided into two main categories; \textit{Static Watermarking} and \textit{Dynamic Watermarking} \cite{li2021survey}.

\emph{Static Watermarking} consists of methods which embed a watermark in a protected model's structure and/or weights. In this case, watermark extraction occurs through static analysis of the protected model's parameters/structure.
Static Watermarking frameworks require white-box access to the suspect model in order to perform watermark extraction \cite{li2021survey}. In cases where the suspect model is only available through an API, or through some other cloud-based or Machine-Learning-as-a-Service (MLaaS) framework, the watermark will not be accessible to the protected model owner. This is a critical disadvantage in modern IP protection use-cases.
Various methods were suggested to determine the origin of a DNN through static analysis (assuming physical access to the model): matrix multiplication on convolutional layers \cite{nagai2018digital}, the usage of passport layers \cite{fan2019extended}, etc.

\emph{Dynamic Watermarking} consists of methods which embed watermarks by training/generating a "key-set" of inputs, and verify the watermark through querying the key-set and observing the suspect model's output.
Dynamic Watermarking methods require black-box access to the suspect model in order to perform watermark extraction/verification. In cases where the suspect model is only available through an API, the watermark will be accessible to the protected model owner, enabling remote watermark extraction \cite{li2021survey}.
Various methods for creating the key-set needed for Dynamic Watermarking were suggested: the use of input perturbations near a decision boundary \cite{merrer2019frontier}, embedding a mask on legitimate input \cite{zhang2018protecting,guo2018embedded,li2020piracy}, using abstract or out-of-distribution input \cite{zhang2018protecting,adi2018weakness}, adding noise to legitimate input \cite{zhang2018protecting}, training an encoder to generate watermarked input \cite{li2019blind}, randomly misclassifying a subset of legitimate input \cite{namba2019robust,jia2021entangled}, and generating perturbed adversarial input to use as a watermark \cite{Zhao2020AFAAF,lukas2021deep}.
Despite these advantages mentioned above, current frameworks are vulnerable to model extraction attacks, along with other strong and adaptive attacks, as demonstrated by \cite{lukas2021sok}.

%% file: hypothesis.tex
\section {Initial Seeds as an Inherent Property}
\label{sec:analysis}

\begin{figure*}
\begin{minipage}{\textwidth}
  \begin{minipage}[b]{0.24\textwidth}
    \centering
    \includegraphics[width=0.6\columnwidth]{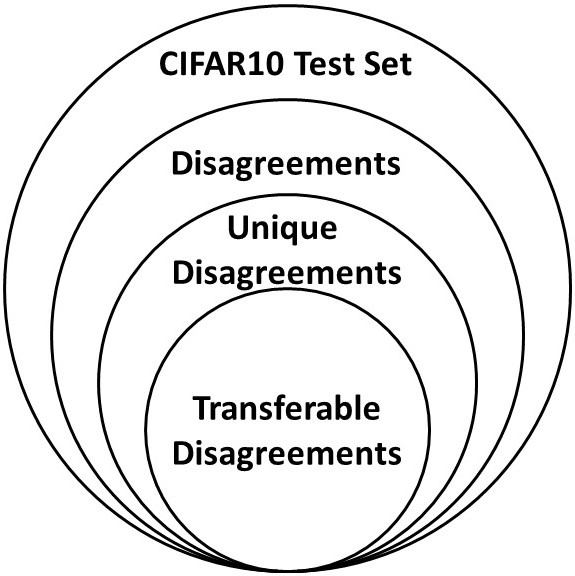}
    \captionof{figure}{The three analyzed subsets of the CIFAR10 test set.}
    \label{fig:subset-visualization}
  \end{minipage}
  \hfill
  \begin{minipage}[b]{0.74\textwidth}
    \centering
    \begin{tabular}{|c|c|c|c|c|}
\hline
                                  & Disagreements & Unique & Transferable &  Confidence \\ \hline
No BIM                            & \textbf{70.4\%}                   & \textbf{9.96\%}                         & 1.57\%                              & 0.61                                \\ \hline
Strategy 1 (Unique) & \textbf{70.4\%}                   & \textbf{9.96\%}                         & 0.58\%                               & 0.97                                \\ \hline
Strategy 2 (Disagreements)        & \textbf{70.4\%}                   & 9.62\%                         & \textbf{1.77\%}                               & 0.97                                \\ \hline
Strategy 3 (Entire Dataset)       & 37.2\%                  & 4.55\%                         & 0.6\%                               & \textbf{0.98}                                \\ \hline
\end{tabular}
      \captionof{table}{Share of disagreements, unique disagreements, and transferable disagreements, as well as transferable confidence scores, when applying BIM according to the predefined strategies.}
      \label{tab:analysis-results}
    \end{minipage}
  \end{minipage}
  \vspace{-1.5em}
\end{figure*}

In this section, we establish the fundamentals behind the key concepts of our proposed method. Specifically, we show that the initial random seed used during model training can be used for attestation of unique behavior in a protected DNN model which transfers to extracted models, and is absent from other non-extracted models trained independently on the training data.

We demonstrate how utilizing such a seed prior to training a DNN model uniquely influences its decision boundary. 
In addition, we demonstrate how extracted models also inherit this boundary. 
Since these extracted models possess similar decision boundaries to the protected model, there exists a subset of inputs which are misclassified by extracted models, and correctly classified by non-extracted models (models that were independently trained on the training data). These inputs can serve as a basis of a watermark for the protected model and its extracted models (models obtained from the protected model via a model extraction attack), attesting to the unique behavior present in their similar decision boundaries.

\subsection{Experimental Setup}
The data-set used in this analysis is the CIFAR10 data-set \cite{cifar10}. CIFAR10 is an image classification data-set commonly used for bench-marking in the CV field. 
The data-set consists of images from ten classes. 
Using training and test sets of 50,000 and 10,000 images, respectively, we trained 20 protected DNN models. 
The architecture used for the models is a convolutional architecture with the following layers:
(1) Input - 32x32x3 image.
(2) Convolutional layer with 32 3x3 filters with ReLU activation, followed by 2x2 Max Pooling.
(3) Convolutional layer with 64 3x3 filters with ReLU activation, followed by 2x2 Max Pooling.
(4) Convolutional layer with 64 3x3 filters with ReLU activation.
(5) Dense layer with 64 neurons with ReLU activation.
(6) Dense layer with 10 neurons with Softmax activation.
(7) Output - 10 Softmax values corresponding to class confidence scores.

Each of the 20 protected models had identical architectures, and were trained on the same data-set using the Adam optimizer \cite{Kingma2014Adam}. The only difference between the 20 training processes was a unique random seed that was uniquely and randomly assigned.

After training the 20 protected models for 10 epochs, we applied the Copycat CNN \cite{copycatcnn} model extraction attack for 15 epochs to each of the protected models. The Copycat CNN attack queries the target model with random data to build a labelled data-set, and then trains a surrogate model with similar architecture on the obtained data-set.
This extraction was performed with a random subset of 50\% of the training data (\textit{i.e.,} 25000 queries were made to the protected model), and produced 20 extracted models, one per protected model.

\subsection{Model Decision Boundary Share Analysis}
The first analysis addresses the decision boundaries of the 20 protected models. 
In particular, three subsets of the CIFAR10 test set were investigated:
\textbf{(1) Disagreements:} Inputs which at least two protected models classify to different classes. 
\textbf{(2) Unique Disagreements:} Disagreements which one protected model misclassifies, and the rest of the protected models correctly classify.
\textbf{(3) Transferable Disagreements:} Unique disagreements which both the protected model and its extracted model misclassify to the same class. Since these misclassifications are unique behavior in the protected model (compared to the other protected models), and are inherited by its extracted model, the transferable disagreements subset can be utilized for creating a watermark key-set for the protected model.
The subsets are visualized in Fig. \ref{fig:subset-visualization}. 
We aimed to determine the proportions of these three subsets in the CIFAR10 test set.

\textit{Results:} Table \ref{tab:analysis-results} displays the results of this analysis. We found that disagreements make up 70.4\% of the data-set. Also, unique disagreements make up 9.96\% of the data-set. Finally, transferable disagreements make up 1.57\% of the data-set, or nearly 16\% of unique disagreements, with an average confidence score of 0.61.
Therefore, we can conclude that random seed initialization uniquely affects a model's decision boundary, producing unique disagreements. In addition, a substantial amount of unique disagreements (approx. 16\%) transfer to extracted models.

\subsection{Model Decision Boundary Confidence \\ Analysis}
The analysis in this section utilizes the Basic Iterative Method (BIM) \cite{kurakin2016adversarial}, an iterative gradient-based method for generating adversarial examples misclassified by DNN models. BIM can be targeted toward a desired target class or untargeted. We chose BIM as a method to increase the robustness of the watermark by exploiting the protected model's unique decision boundary to produce examples that it misclassifies with a higher confidence score. We note that any method which produces disagreements with stronger confidence scores can also be utilized.

This analysis investigates the impact of BIM on transferability, \textit{i.e.,} the size and confidence scores of the transferable disagreements subset. A higher confidence score indicates a more robust watermark with a higher likelihood of surviving model extraction.

Three strategies of utilizing BIM were analyzed:
\textbf{Strategy 1:} Applying BIM to the unique disagreements subset.
\textbf{Strategy 2:} Applying BIM to the disagreements subset.
\textbf{Strategy 3:} Applying BIM to the entire CIFAR10 test set.

Table \ref{tab:analysis-results} displays the results of this analysis. These results demonstrate that BIM increases the robustness of the watermark key-set by improving the confidence scores of the transferable disagreements. It can be seen that the most effective strategy of applying BIM is Strategy 2 (applying BIM to the disagreements subset). In this case, the size of the transferable disagreements subset remained nearly the same, while the average confidence of the transferable disagreements increased from 0.61 to 0.97.
Therefore, we can conclude that BIM improves the confidence scores of the transferable disagreements, increasing watermark robustness on the protected and extracted models.

%% file: method.tex
\section{Threat Model \& RAW Framework}
\label{sec:methodology}
In this section, we describe the threat model and methodology for our proposed framework, the RAW Framework.

\subsection{Threat Model}
As part of the domain of IP protection for a DNN model, we describe the assumptions made regarding potential attackers. An attacker is assumed to copy/extract the protected model via an API (model extraction).
The attacker is assumed to have black-box (API) access to the protected model, including access to the protected model's confidence scores in response to a query. We also assume partial access to input from the protected model's target domain.
The attacker's knowledge of the watermark is assumed to fall into one of two categories:
(1) \textit{Naive Attacker} - The attacker doesn't know that the protected model is watermarked. The attacker performs a model extraction attack to obtain the model.
(2) \textit{Informed Attacker} - The attacker knows that the protected model is watermarked. However, the attacker is assumed to not know which attacks the watermark is robust against, or which inputs are in the watermark key-set. The attacker is assumed to utilize model extraction and blurring attacks to obtain the model and blur unique behavior attested to by the watermark (e.g., model manipulation attacks (such as weight pruning) and input permutation).
The model owner/protector is assumed to have white-box access to the protected model, as well as black-box access to suspect models which they believe may be extracted models, including access to the suspect model's confidence scores in response to a query.

\subsection{RAW Framework}

We propose the Robust Adaptive Watermarking (RAW) Framework, a novel adaptive framework, for providing robust IP protection to a protected DNN model. Given a protected model and set of attacks, the RAW Framework generates a watermark key-set and verification model for the protected model from input provided to the framework. The protected model is provided non-watermarked, and no alterations are made to the model in the watermark generation process. The attacks are utilized locally by the model owner to simulate an attacker extracting the protected model.
A set of non-extracted models, which are trained independently of the protected model and initialized with different initial seeds, are provided as input for use as a control group, in order to ensure that the verification model attests to watermarked behavior exclusively on the protected model and extracted models. In addition, input data is provided for generating the watermark key-set.

\begin{figure}[h!]
    \centering
    \vspace{-0.5em}
    \includegraphics[width=\columnwidth]{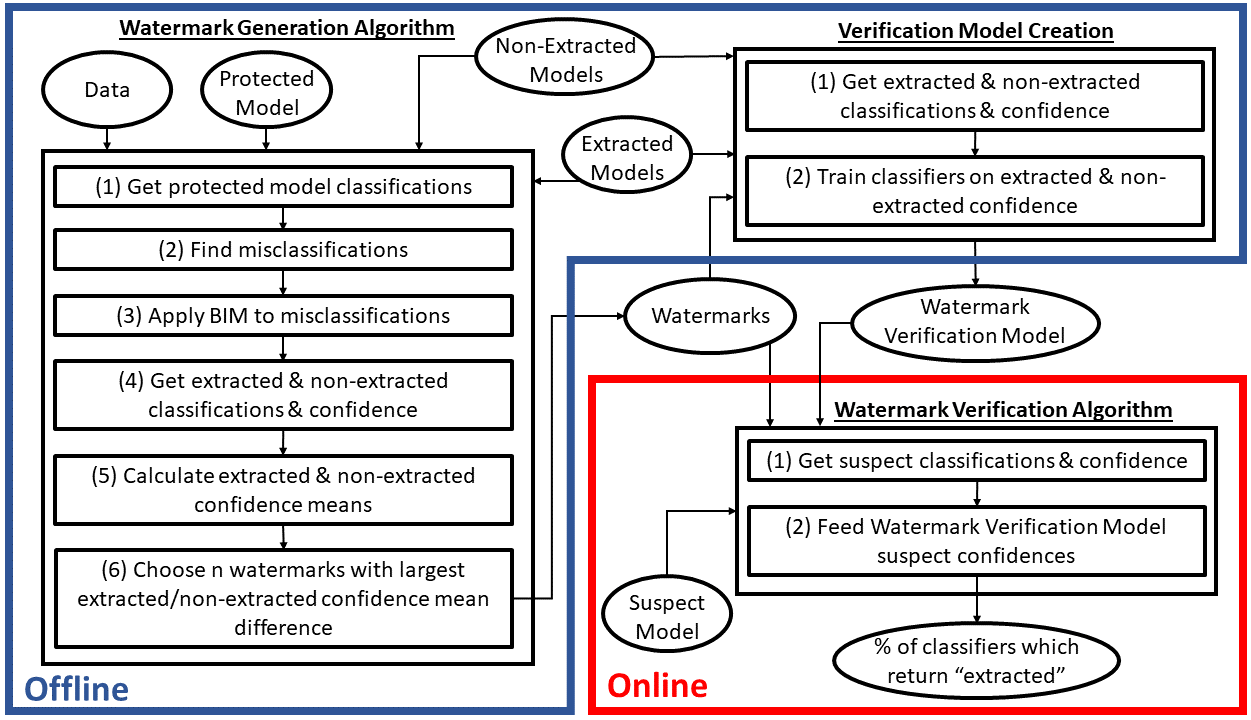}
    \caption{High-level architecture of the key-set generation, verification model creation, and watermark verification algorithms used in the RAW Framework.}
    \label{high-level-key-generation}
\end{figure}

The RAW Framework consists of two stages; an offline and an online stage. The offline stage involves (1) generating the watermark key-set and (2) creating/training the verification model. The online stage involves utilizing the watermark key-set and verification model to determine if a suspect model displays watermarked behavior based on its query responses to the watermark key-set.
Fig. \ref{high-level-key-generation} displays a high-level visualization of the framework's steps.

After generating the watermark key-set and verification model with a given set of attacks, the model protector has the ability to update the key-set and verification model in the future in response to new attacks. This is done by repeating the offline stage of the RAW Framework and adding models extracted with the new attacks to the input of the framework. This results in a new watermark key-set and verification model which accounts for the new attacks that weren't considered in the past.

\subsubsection{Watermark Key-Set Creation Algorithm (Offline)}

The watermark key-set generation algorithm is as follows:
\textbf{Input:} Protected model, collection of extracted models (each initialized with a unique seed), collection of non-extracted models (each initialized with a unique seed), input from the model's target domain.
\textbf{(1)} Feed the protected model the input data, and obtain the protected model's classifications.
\textbf{(2)} Find \textit{misclassifications}, inputs which the protected model misclassifies.
\textbf{(3)} Obtain \textit{misclassifications*} - increase the confidence of the \textit{misclassifications} subset by applying BIM to \textit{misclassifications} for 20 iterations, targeted to the protected model's predictions.
\textbf{(4)} For each extracted and non-extracted model, feed the model \textit{misclassifications*} and obtain the model's confidence scores on \textit{misclassifications*}.
\textbf{(5)} For each input $x$ in \textit{misclassifications*}, calculate $confidence_{x, e}$ and $confidence_{x, ne}$, the mean confidence scores for the protected model's predicted class obtained by the extracted models and non-extracted models, respectively.
\textbf{(6)} Designate/sort the $n$ inputs from \textit{misclassifications*} with the largest difference $|confidence_{x, e} - confidence_{x, ne}|$ as the watermark key-set ($n$ is determined by the protected model owner).
\textbf{Output:} Watermark key-set, each watermark labelled to the protected model's predicted class.

\subsubsection{Verification Model Generation (Offline)}

\begin{figure}[h!]
    \centering
    \vspace{-1em}
    \includegraphics[width=0.9\columnwidth]{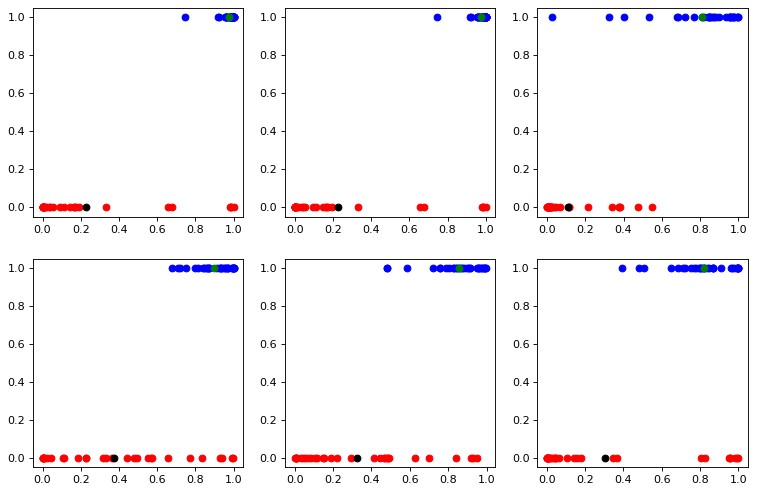}
    \caption{Confidence scores (displayed on horizontal axis) of extracted (blue) and non-extracted (red) models on six watermarks from the watermark key-set. The average extracted and non-extracted confidence scores are marked in green and black, respectively.}
    \label{fig:confidence_distributions}
\end{figure}

The watermarks in the watermark key-set are chosen due to their: (1) demonstration of unique behavior in the protected model, (2) transferability to extracted models, and (3) optimal difference in mean confidence scores between extracted and non-extracted models. Fig. \ref{fig:confidence_distributions} displays the confidence scores of extracted and non-extracted models (including the average scores) on six watermarks in the watermark key-set. Confidence scores on the entire key-set are displayed in Fig. \ref{fig:keyset_confidence_distributions} in the appendix. It can be seen that the extracted confidence values tend to be higher than the non-extracted confidence values. Therefore, a Logistic Regression classifier trained on extracted and non-extracted confidence values for a particular watermark would learn a confidence score threshold for distinguishing between "extracted"and "not extracted" models. In addition, the average confidences of the extracted and non-extracted models tend to have a distinct margin between them. Therefore, a Gaussian Naive Bayes classifier could also be an effective discriminatory tool for classifying a model as "extracted" or "non-extracted" based on its confidence score on a watermark.

Based on the observations of the confidence score distributions, the verification model is trained/created as follows:
\textbf{Input:} Collection of extracted models (same models from key-set generation), collection of non-extracted models (same models from key-set generation), watermark key-set, watermark labels.
\textbf{(1)} For each model (extracted and non-extracted), feed the model the watermark key-set and obtain the model's confidence scores pertaining to the watermark labels.
\textbf{(2)} For each watermark $wm$, train a classifier (Logistic Regression or Gaussian Naive Bayes) on the extracted and non-extracted models' confidence scores for $wm$. The classifier is a binary classifier with output of "extracted" or "not extracted" given a model's confidence on the watermark.
\textbf{Output:} Return the watermark key-set and labels, along with each watermark's respective classifier.
We note that since verification model creation occurs during the offline stage, the protector has access to the same extracted and non-extracted models used to generate the watermark key-set.

\subsubsection{Watermark Verification Algorithm (Online)}
Watermark verification is performed by feeding a suspect model's watermark key-set query responses to the verification model, which outputs a confidence score of the suspect model being watermarked. Since the model owner/protector performs both the offline and online stages of the RAW Framework, they have access to the original verification model and watermark key-set.
The detailed steps of the watermark verification algorithm are as follows:
\textbf{Input:} Suspect model, verification model (created during the offline stage), watermark key-set, watermark labels.
\textbf{(1)} Feed the suspect model the watermark key-set and obtain the model's confidence scores pertaining to the watermark labels.
\textbf{(2)} For each watermark $wm$, feed the suspect model's confidence for $wm$ to its respective classifier in the verification model, and receive a classification of "extracted" or "not extracted" for each classifier.
\textbf{Output:} Return the percentage of watermarks where the suspect model's confidence score is classified as "extracted" by the watermark's respective classifier. This is the confidence score of the verification model that the suspect model is extracted from the protected model.

%% file: evaluation.tex
\section{Evaluation}
\label{sec:evaluation}
In this section, we evaluate the performance of the RAW Framework against two types of attackers.
The first attacker type is a \emph{naive attacker} utilizing model extraction attacks to obtain the protected model, and the second type is an \emph{informed attacker} who: (1) obtains the protected model via a model extraction attack, and then (2) applies a blurring method to the extracted model in order to remove/obscure watermarked behavior. In addition, we evaluate the generality of the RAW Framework to extracted models with (1) the same architecture but different structure, and (2) different architecture than the protected model.
The RAW Framework takes advantage of the unique watermarked behavior inherent to the protected model's decision boundary, and expressed via the model's confidence scores for its misclassifications. This behavior is dependent on, and a result of, the initial random seed utilized during the protected model's training. In addition, this behavior is inherited by extracted models, and by identifying the existence of this behavior, the RAW Framework is capable of identifying when a suspect model is extracted from the protected model.

\subsection{Robustness Against a Naive Attacker}
The first case we evaluate is a naive attacker who obtains the protected model through utilizing a model extraction attack to train a surrogate model with the same architecture and structure as the protected model. 
We evaluate the performance (ROC/AUC and TPR/FPR) of the verification model at detecting extracted models, given a suspect model's confidence scores on the watermark key-set.

\textbf{Experimental Setup:} Initially, we trained a ResNet model (the "protected model") on the CIFAR10 data-set, initialized with a unique random seed during training.
The model's architecture is a Wide ResNet \cite{wideresnet} architecture, consisting of a 3x3 convolutional layer, followed by three convolutional blocks, and ending with a batch-norm layer and a linear transformation layer which outputs predictions for each of the 10 CIFAR10 classes. Each convolutional block consists of three five-layer blocks (batch-norm, 3x3 convolution, dropout, batch-norm and 3x3 convolution), with each block in the last two convolutional blocks also containing a 1x1 convolutional skip layer.

The model extraction attacks utilized in this evaluation were Retraining \cite{DBLP:journals/corr/TramerZJRR16}, Distillation \cite{https://doi.org/10.48550/arxiv.1503.02531}, and Transfer Learning \cite{torrey2010transfer}. These three attacks were mentioned in \cite{lukas2021sok} for evaluating various watermarking methods' robustness to extraction attacks, and nearly every method was found to be vulnerable to Retraining and/or Transfer Learning.

The evaluation process was as follows:
\textbf{(1)} 30 models are provided to the RAW Framework as extracted models. They are extracted using either one or two different "seen" extraction attacks (attacks which the protector has access to for watermark key-set and verification model generation). If using more than one extraction attack, then an equal number of models are provided using each attack (totaling to 30).
\textbf{(2)} 30 models are provided to the RAW Framework as non-extracted models. They are independently trained and have either one of two Wide ResNet structures or a DenseNet \cite{Huang2016DenseNet} architecture. The number of provided models with each architecture/structure was equal.
\textbf{(3)} The protected model, the extracted models from (1), the non-extracted models from (2), and input from the CIFAR10 data-set are provided to the RAW Framework as a "training set" to generate the watermark key-set and verification model.
\textbf{(4)} 30 models (15 extracted + 15 non-extracted) are used as a "test set" to evaluate the watermark key-set and verification model. The extracted models were obtained using an "unseen" extraction attack not utilized in (1). The non-extracted models possess the same architectures as in the training set, and were initialized with unique random seeds prior to training.
This process will be referred to in the following evaluations as the "RAW evaluation process".

The "RAW evaluation process" was performed 20 times for this evaluation, with a new protected model trained with a new seed each time. The evaluation was performed twice, using two types of classifiers in the verification model: Logistic Regression and Gaussian Naive Bayes. The evaluated attack configurations, and aggregated results are reported below.

\begin{figure}[]
\begin{minipage}{1.0\columnwidth}
  \begin{minipage}[b]{0.49\columnwidth}
    \includegraphics[width=1.0\columnwidth]{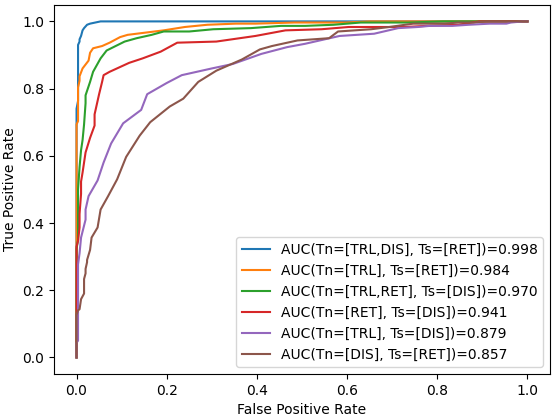}
    \vspace{-1.0em}
  \end{minipage}
  \begin{minipage}[b]{0.49\columnwidth}
    \includegraphics[width=1.0\columnwidth]{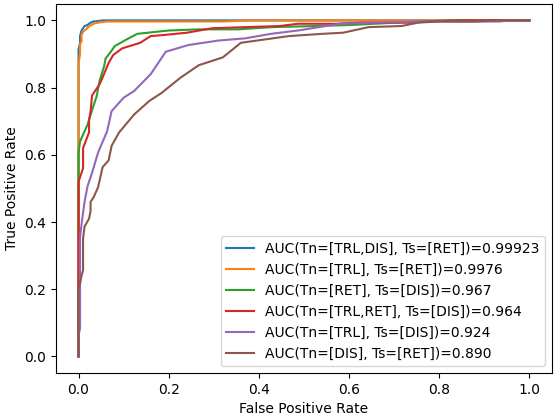}
    \vspace{-1.0em}
  \end{minipage}
  \caption{ROC/AUC values when detecting a naive attacker using Logistic Regression (left) and Gaussian Naive Bayes (right). (Tn = training set, Ts = test set, TRL = Transfer Learning \cite{torrey2010transfer}, RET = Retraining \cite{DBLP:journals/corr/TramerZJRR16}, DIS = Distillation \cite{https://doi.org/10.48550/arxiv.1503.02531})}
    \label{fig:naive-evaluation-roc-auc}
 \end{minipage}
\end{figure}

\begin{table}
\centering
\resizebox{\columnwidth}{!}{%
\begin{tabular}{|cc|cc|cc|cc|}
\hline
\multicolumn{2}{|c|}{Configuration}   & \multicolumn{2}{c|}{TPR(FPR=0)}  & \multicolumn{2}{c|}{FPR(TPR=1)}  & \multicolumn{2}{c|}{AUC}           \\ \hline
\multicolumn{1}{|c|}{Train}    & Test & \multicolumn{1}{c|}{LR}   & GNB  & \multicolumn{1}{c|}{LR}   & GNB  & \multicolumn{1}{c|}{LR}    & GNB   \\ \hline
\multicolumn{1}{|c|}{TRL, DIS} & RET  & \multicolumn{1}{c|}{0.74} & 0.91 & \multicolumn{1}{c|}{0.05} & 0.05 & \multicolumn{1}{c|}{0.998} & 0.999 \\ \hline
\multicolumn{1}{|c|}{TRL}      & RET  & \multicolumn{1}{c|}{0.69} & 0.87 & \multicolumn{1}{c|}{0.65} & 0.38 & \multicolumn{1}{c|}{0.984} & 0.997 \\ \hline
\multicolumn{1}{|c|}{DIS}      & RET  & \multicolumn{1}{c|}{0.12} & 0.21 & \multicolumn{1}{c|}{0.91} & 0.92 & \multicolumn{1}{c|}{0.857} & 0.89  \\ \hline
\multicolumn{1}{|c|}{TRL, RET} & DIS  & \multicolumn{1}{c|}{0.32} & 0.52 & \multicolumn{1}{c|}{0.81} & 0.84 & \multicolumn{1}{c|}{0.97}  & 0.964 \\ \hline
\multicolumn{1}{|c|}{RET}      & DIS  & \multicolumn{1}{c|}{0.33} & 0.61 & \multicolumn{1}{c|}{0.89} & 0.83 & \multicolumn{1}{c|}{0.941} & 0.967 \\ \hline
\multicolumn{1}{|c|}{TRL}      & DIS  & \multicolumn{1}{c|}{0.04} & 0.06 & \multicolumn{1}{c|}{0.99} & 0.95 & \multicolumn{1}{c|}{0.879} & 0.924 \\ \hline
\end{tabular}
}
      \captionof{table}{Constrained TPR/FPR values and AUC values when detecting a naive attacker. (LR = Logistic Regression, GNB = Gaussian Naive Bayes, TRL = Transfer Learning \cite{torrey2010transfer}, RET = Retraining \cite{DBLP:journals/corr/TramerZJRR16}, DIS = Distillation \cite{https://doi.org/10.48550/arxiv.1503.02531}).}
      \label{tab:tpr-fpr-values-table-naive}
      \vspace{-1.5em}
\end{table}

\textbf{Results:} Fig. \ref{fig:naive-evaluation-roc-auc} shows the ROC/AUC values for each evaluated configuration of extraction attacks.
On every attack configuration, the RAW watermark verification model obtained outstanding (>0.9) or near-outstanding AUC values, with both verification models (Logistic Regression and Gaussian Naive Bayes).
The highest performing configurations were able to detect extracted models from the Retraining attack with AUC values over 0.99. This indicates that this attack produces a model with behavior close to the behavior attested to by these configurations' verification model. 
Other attacks can succeed to obtain the protected model's functionality while obtaining more obscure decision boundaries. Identifying these boundaries requires utilizing a diverse range of attacks in the RAW Framework in order to generate a more robust watermark key-set and verification model.
It can be seen that utilizing a more diverse range of attacks in the RAW Framework produces a more robust watermark key-set, reflected in higher AUC values for nearly every configuration.

In addition, we also evaluated the accuracy of the verification model according to two policies: (1) a False-Positive Rate (FPR) of 0, \textit{i.e.,} no non-extracted model is misclassified as extracted, and (2) a True-Positive Rate of 1, \textit{i.e.,} every extracted model is classified as extracted. These results are reported in Table \ref{tab:tpr-fpr-values-table-naive}. We can see that when detecting Retraining, the verification model obtains a TPR of 0.91 according to policy (1), and an FPR of 0.05 according to policy (2), when using Gaussian Naive Bayes.

\begin{figure*}
\begin{subfigure}{0.32\textwidth}
\centering
    \includegraphics[width=\textwidth]{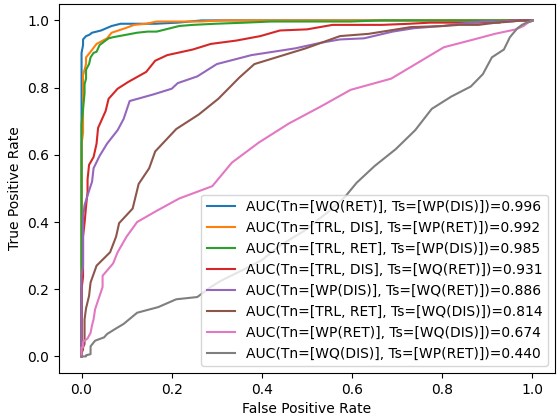}
    \caption{Robustness against an informed attacker.}
    \label{fig:sophisticated-evaluation-roc-auc}
\end{subfigure}
\hfill
\begin{subfigure}{0.32\textwidth}
    \centering
    \includegraphics[width=\textwidth]{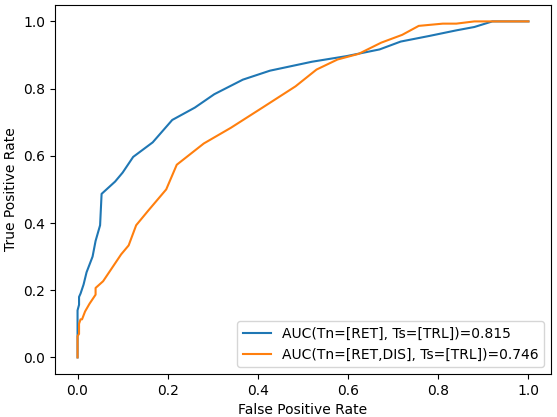}
    \caption{Generality to different model structures.}
    \label{fig:cross-structure-evaluation-roc-auc}
\end{subfigure}
\hfill
\begin{subfigure}{0.32\textwidth}
    \centering
    \includegraphics[width=\textwidth]{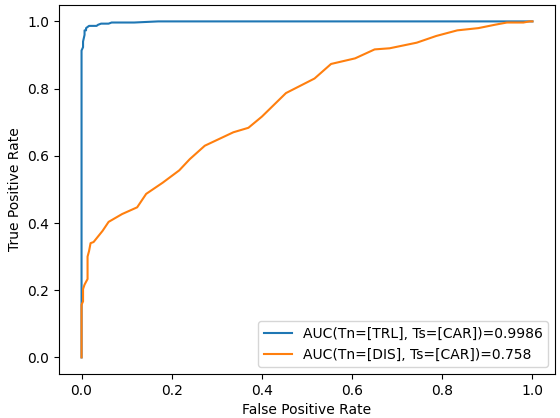}
    \caption{Generality to different model architectures.}
    \label{fig:cross-architecture-evaluation-roc-auc}
\end{subfigure}
\vspace{-0.5em}
\caption{ROC/AUC values against an informed attacker (left), as well as against different model structures (center) and architectures (right). Tn = training set, Ts = test set, WP = Weight Pruning \cite{https://doi.org/10.48550/arxiv.1710.01878}, WQ = Weight Quantization \cite{https://doi.org/10.48550/arxiv.1609.07061}, RET = Retraining \cite{DBLP:journals/corr/TramerZJRR16}, DIS = Distillation \cite{https://doi.org/10.48550/arxiv.1503.02531}, TRL = Transfer Learning \cite{torrey2010transfer}, CAR = Cross-Architecture Retraining \cite{DBLP:journals/corr/TramerZJRR16}).}
\vspace{-1.5em}
\end{figure*}

\begin{table}[]
\centering
\resizebox{\columnwidth}{!}{%
\begin{tabular}{|c|c|c|c|c|}
\hline
Train    & Test    & TPR(FPR=0) & FPR(TPR=1) & AUC \\ \hline
WQ(RET)  & WP(DIS) & 0.90        & 0.27      & 0.996  \\ \hline
TRL, DIS & WP(RET) & 0.63        & 0.31      & 0.992  \\ \hline
TRL, RET & WP(DIS) & 0.69        & 0.66      & 0.985  \\ \hline
WP(DIS)  & WQ(RET) & 0.24        & 0.96      & 0.886  \\ \hline
TRL, DIS & WQ(RET) & 0.2        & 0.96      & 0.931  \\ \hline
TRL, RET & WQ(DIS) & 0.01        & 0.97       & 0.814 \\ \hline
WP(RET)  & WQ(DIS) & 0.03        & 1.0      & 0.674  \\ \hline
WQ(DIS)  & WP(RET) & 0.0        & 1.0      & 0.44  \\ \hline
TRL      & CAR     & 0.91       & 0.17      & 0.999 \\ \hline
DIS      & CAR     & 0.16       & 0.99      & 0.758 \\ \hline
\end{tabular}
}
\vspace{-0.5em}
\captionof{table}{Constrained TPR/FPR values and AUC values when detecting (1) an informed attacker, and (2) extracted models with different architectures. (WP = Weight Pruning \cite{https://doi.org/10.48550/arxiv.1710.01878}, WQ = Weight Quantization \cite{https://doi.org/10.48550/arxiv.1609.07061}, TRL = Transfer Learning \cite{torrey2010transfer}, RET = Retraining \cite{DBLP:journals/corr/TramerZJRR16}, DIS = Distillation \cite{https://doi.org/10.48550/arxiv.1503.02531}, CAR = Cross-Architecture Retraining \cite{DBLP:journals/corr/TramerZJRR16}).}
      \label{tab:tpr-fpr-values-table}
      \vspace{-1.5em}
\end{table}

\subsection{Robustness Against an Informed Attacker}
In this next case, we evaluate an informed attacker who: (1) obtains the protected model through utilizing a model extraction attack, and (2) performs a watermark blurring strategy in order to remove watermarked behavior from the extracted model. As in the previous evaluation, we evaluate the ROC/AUC and TPR/FPR of the verification model.

\textbf{Experimental Setup:} We utilized the same experimental setup described previously.
The model extraction attacks utilized in this evaluation were: Retraining \cite{DBLP:journals/corr/TramerZJRR16}, Distillation \cite{https://doi.org/10.48550/arxiv.1503.02531}, and Transfer Learning \cite{torrey2010transfer}.
In addition, in order to simulate an informed attacker, we utilized two blurring strategies: Weight Pruning \cite{https://doi.org/10.48550/arxiv.1710.01878} with 0.5 sparsity, and Weight Quantization \cite{https://doi.org/10.48550/arxiv.1609.07061}.
The "RAW evaluation process" was repeated 20 times in this evaluation, with a new protected model each time, using Logistic Regression in the watermark verification model. The attack configurations utilized in the training set (used to generate the watermark key-set and train the verification model) were either: (1) two "naive" attacks, or (2) a naive attack with a blurring method applied to the resulting models. The extracted models in the test set consisted of models extracted with one of the naive extraction attacks with blurring applied afterward. The evaluated attack configurations and aggregated results are reported below.

\textbf{Results:} Fig. \ref{fig:sophisticated-evaluation-roc-auc} shows the ROC/AUC values for each evaluated configuration of "seen" and "unseen" extraction attacks and blurring methods. It can be seen that configurations where the models obtained with the Retraining attack are blurred with Weight Quantization, and the models obtained with the Distillation attack are blurred with Weight Pruning, achieve the highest AUC values when used to detect one-another (over 0.88). This indicates that the models resulting from these configurations inherit similar watermarked behavior from the protected model, which the verification model attests to. In addition, there is a configuration for every evaluated informed extraction attack enabling detection with an AUC of over 0.81, indicating that informed usage of the RAW Framework by the protected model owner provides robust capabilities to detect extracted models, including those which undergo blurring methods to remove/obscure watermarked behavior originating from the unique random seed used during model training.
Table \ref{tab:tpr-fpr-values-table} displays the TPR/FPR values according to the two previously described policies. It can be seen that for two of the four evaluated test configurations, there is a train configuration enabling detection of extracted models with a TPR over 0.6 when constraining the FPR to 0, and an FPR of under 0.31 when constraining the TPR to 1.

\subsection{Generalization to Differing Model Structure}
In this section, we evaluate the generalization of the RAW Framework to detecting extracted models with the same architecture, but different structure than the protected model. In particular, we evaluate the performance of the verification model when detecting models extracted with Transfer Learning, where the surrogate model has the ResNet34 architecture \cite{He2015resnet34}.

\textbf{Experimental Setup:}
We utilized the same experimental setup described previously.
The model extraction attacks utilized in this evaluation were Retraining \cite{DBLP:journals/corr/TramerZJRR16}, Distillation \cite{https://doi.org/10.48550/arxiv.1503.02531}, and Transfer Learning \cite{torrey2010transfer}.
The "RAW evaluation process" was repeated 20 times in this evaluation, with a new protected model each time, using Logistic Regression in the watermark verification model. The extracted models in the test set consisted of models extracted with Transfer Learning. The evaluated attack configurations and aggregated results are reported below.

\textbf{Results:}
Fig. \ref{fig:cross-structure-evaluation-roc-auc} shows the ROC/AUC values for the evaluated configurations. It can be seen that utilizing Retraining allows the protected model owner to detect models extracted with Transfer Learning with an AUC value of over 0.81. Based on our previous findings, utilizing a more robust configuration of attacks in the RAW Framework would help improve the generality of the watermark key-set and verification model and further improve the watermark verification performance for this extraction attack. This demonstrates the importance of a model protector carefully considering the model extraction attacks they are defending against, and utilizing a configuration of attacks to produce a verification model which best addresses these attacks.

\subsection{Generalization to Differing Model Architecture}
In this section, we evaluate the generalization of the RAW Framework to detecting extracted models with different architectures than the protected model. In particular, we evaluate the performance of the verification model when detecting models extracted with Cross-Architecture Retraining, where the surrogate model has the DenseNet architecture.

\textbf{Experimental Setup:}
We utilized the same experimental setup described previously.
The model extraction attacks utilized in this evaluation were: Cross-Architecture Retraining \cite{DBLP:journals/corr/TramerZJRR16}, Distillation \cite{https://doi.org/10.48550/arxiv.1503.02531}, and Transfer Learning \cite{torrey2010transfer}.
The "RAW evaluation process" was repeated 20 times in this evaluation, with a new protected model each time, using Logistic Regression in the watermark verification model. The extracted models in the test set consisted of models extracted with Cross-Architecture Retraining. The evaluated attack configurations and aggregated results are reported below.

\textbf{Results:}
Fig. \ref{fig:cross-architecture-evaluation-roc-auc} shows the ROC/AUC values for the evaluated configurations. It can be seen that utilizing Transfer Learning allows the protected model owner to detect models extracted with Cross-Architecture Retraining with an AUC value of over 0.99. This demonstrates that the verification model has the ability to detect extracted models with different architecture than the protected model.
Table \ref{tab:tpr-fpr-values-table} displays the TPR/FPR values according to the two previously described policies. It can be seen that using Transfer Learning to generate the watermark key-set and verification model enables detection of models extracted with Cross-Architecture Retraining with a 0.91 TPR when the FPR is 0, and an FPR of 0.17 when the TPR is 1.

%% file: conclusion.tex
\section{Conclusions \& Future Work}
\label{sec:conclusion}

In this paper, we demonstrate that DNNs possess unique behavior due to an initial random seed, which can be used to watermark the model and distinguish between extracted models and non-extracted models. We show that this behavior is present in the confidence scores of the model's misclassifications, and is enhanced when utilizing BIM to produce stronger misclassification confidence. We utilize this phenomenon to produce a watermarking framework which can help detect models extracted from a protected model. We propose the RAW Framework, which when given a protected model, extracted models, and non-extracted models, produces a watermark key-set and verification model that attests to the watermarked behavior inherited by extracted models and not present in non-extracted models. We demonstrate the RAW Framework's robustness to naive attackers who apply a model extraction attack, and to informed attackers who obtain an extracted model and apply a blurring method to obscure watermarked behavior. In addition, we demonstrate the generality of the RAW Framework to different model structures (with the same architecture) and different model architectures.

For future research directions, we propose evaluating the generality of the insights found in this study to models trained on additional CV data-sets, such as ImageNet \cite{ImageNet}.

%% file: appendix.tex
\onecolumn
\section{Appendix}
\label{sec:appendix}

\begin{figure}[h!]
    \centering
    \includegraphics[width=1.0\linewidth]{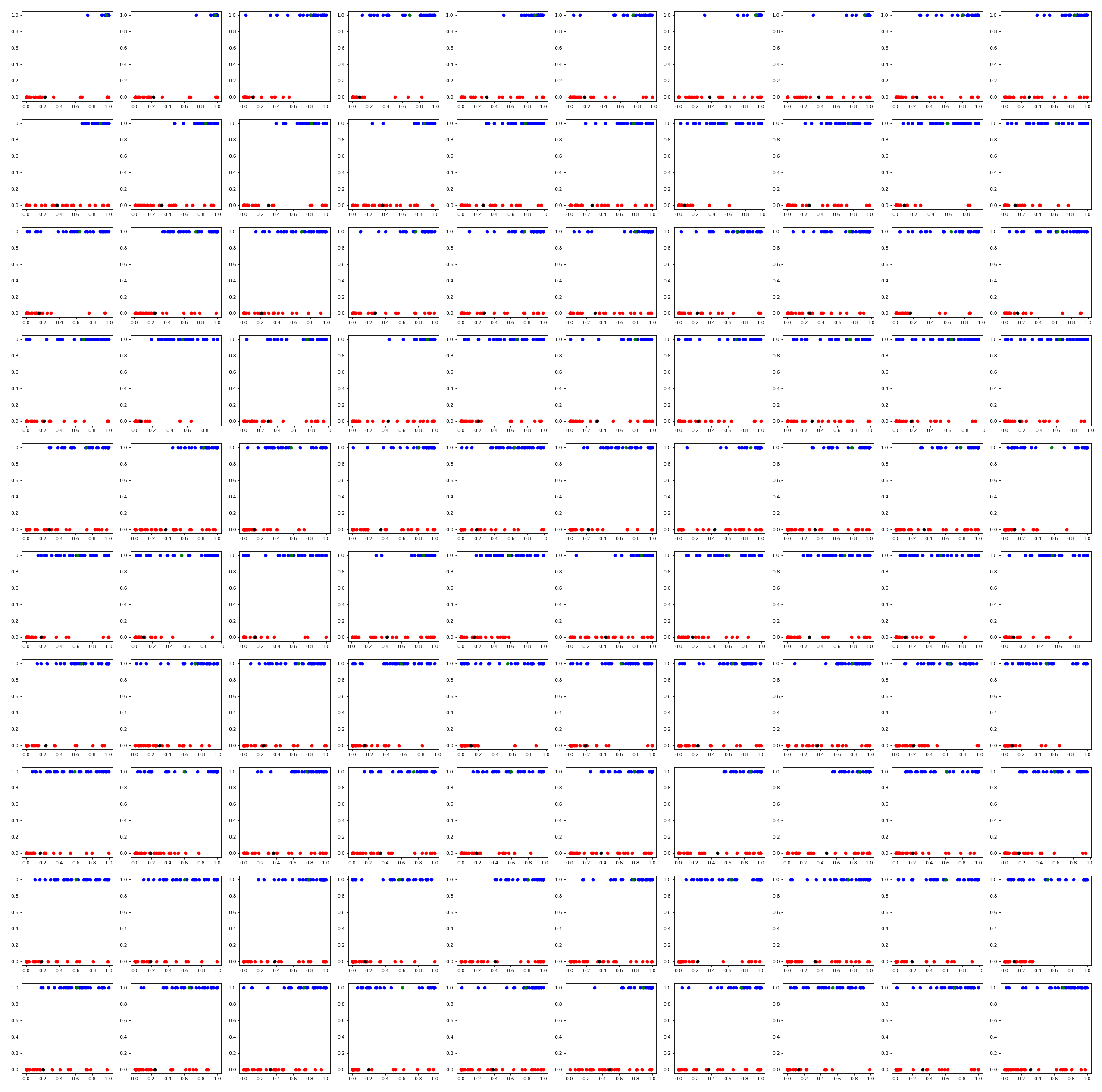}
    \caption{Confidence scores (displayed on horizontal axis) of extracted (blue) and non-extracted (red) models on the entire watermark key-set. The average extracted and non-extracted confidence scores are marked in green and black, respectively.}
    \label{fig:keyset_confidence_distributions}
\end{figure}